\documentclass[runningheads]{llncs}

\usepackage[T1]{fontenc}
\usepackage{graphicx}
\usepackage{color}
\usepackage{booktabs}
\usepackage{array}
\usepackage{tabularx}
\usepackage{makecell}
\usepackage{amsmath,amssymb}

\usepackage{cite}
\usepackage[colorlinks,urlcolor=blue,citecolor=blue,linkcolor=blue]{hyperref}

\begin{document}
	\title{VISP: Volatility Informed Stochastic Projection for Adaptive Regularization}

	\author{Tanvir Islam}
	\institute{Okta, Bellevue, WA, USA\\
	\email{tanvir.islam@okta.com}}

	\authorrunning{T. Islam}

	\maketitle


	\begin{abstract}

		We propose VISP: Volatility Informed Stochastic Projection, an adaptive regularization method that leverages gradient volatility to guide stochastic noise injection in deep neural networks. Unlike conventional techniques that apply uniform noise or fixed dropout rates, VISP dynamically computes volatility from gradient statistics and uses it to scale a stochastic projection matrix. This mechanism selectively regularizes inputs and hidden nodes that exhibit higher gradient volatility while preserving stable representations, thereby mitigating overfitting. Extensive experiments on MNIST, CIFAR-10, and SVHN demonstrate that VISP consistently improves generalization performance over baseline models and fixed-noise alternatives. In addition, detailed analyses of the evolution of volatility, the spectral properties of the projection matrix, and activation distributions reveal that VISP not only stabilizes the internal dynamics of the network but also fosters a more robust feature representation.


	\end{abstract}

	\section{Introduction}
	
	Deep neural networks have achieved remarkable success across a wide spectrum of applications, ranging from computer vision and natural language processing to reasoning agents \cite{Chaplia2024ServerlessAA,Shen2023HuggingGPTSA, Badue2019SelfDrivingCA, Zheng2023JudgingLW}. This is largely due to their ability to learn hierarchical representations from vast amounts of data, enabling them to capture intricate patterns and relationships. However, the capability that makes them powerful also renders them susceptible to overfitting, a phenomenon where the model learns the training data too well, leading to poor generalization on unseen examples. To mitigate this issue, a plethora of regularization techniques have been developed, including noise ingestion \cite{bishop1995training}, weight-based regularization \cite{Smith2018ADA,Louizos2017LearningSN}, dropout \cite{Srivastava2014DropoutAS}, batch normalization \cite{Ioffe2015BatchNA}, various forms of data augmentation \cite{Shorten2019ASO, Yun2019CutMixRS}, among others \cite{Santos2022AvoidingOA}. These techniques introduce various forms of noise or constraints to the learning process, thereby promoting robustness and improving generalization.

	Despite the effectiveness of existing regularization strategies, many of them rely on fixed hyperparameters that are often chosen through extensive and computationally expensive trial-and-error or heuristic approaches. These methods typically employ fixed or uniformly applied regularization strategies, which may not fully exploit the underlying dynamics of the network during training. The optimal regularization strength can vary significantly depending on the dataset, network architecture, and training dynamics. Moreover, during the training process, the importance of regularizing different parts of the network or in different directions in the parameter space might evolve. This necessitates adaptive regularization techniques that can dynamically adjust their strength based on the current state of the learning process.

	In this work, we introduce Volatility Informed Stochastic Projection (VISP), a novel adaptive regularization method designed to harness the inherent volatility of gradients to inform the injection of noise into the network. Rather than applying fixed noise levels, VISP computes a per-feature (or per-channel) volatility measure based on the running statistics of gradient magnitudes and variances. This volatility measure is then used to scale a stochastic projection matrix, which perturbs the activations of the network in a data-dependent manner. In essence, VISP selectively regularizes the parts of the network that are more volatile—those exhibiting higher gradient variability—thereby promoting the learning of robust and stable representations. In summary, the core contribution of this work is a novel regularization approach that dynamically adjusts its strength based on the network's internal state.

	\section{Related Work}
	Regularization is a crucial aspect of training deep neural networks to prevent overfitting and improve generalization. Among the various regularization strategies, methods that inject noise into the training process have shown significant promise. These techniques introduce stochasticity into the network's inputs, weights, activations, or gradients, effectively training an ensemble of models implicitly and encouraging the learning of more robust representation.

	Adding noise directly to the input data is one of the earliest forms of noise-based regularization \cite{zur2009noise,matsuoka1992noise}. Bishop \cite{bishop1995training} has demonstrated that training a neural network with small additive Gaussian noise on the input is analytically equivalent to Tikhonov regularization, which penalizes large weights and promotes smoother mappings. This approach encourages the model to be less sensitive to small perturbations in the input, leading to better generalization on unseen data. Data augmentation techniques, which involve applying random transformations to the training data, can also be viewed as a form of input noise injection, as highlighted in works like CutMix \cite{Yun2019CutMixRS}, which mixes and combines different training examples.

	 Another line of work involves injecting noise into the weights of the neural network during training \cite{an1996effects}. This can be achieved by adding random noise to the weight updates or directly to the weight values. Such techniques encourage exploration of the weight space and can prevent the network from settling into sharp minima in the loss landscape, which are often associated with poor generalization. The idea is that by training with noisy weights, the network becomes more resilient to variations in its parameters.

	Similarly, injecting noise into the hidden units has proven to be highly effective \cite{You2018AdversarialNL,noh2017regularizing}. This typically involves adding random values, often drawn from a specific distribution (e.g., Gaussian), to the hidden layer outputs. The goal is to make the network more robust to small perturbations in its internal representations and improve generalization. Dropout \cite{Srivastava2014DropoutAS} is another prominent stochastic regularization technique that operates on the input and hidden units. In dropout, during training, a subset of neurons is randomly "dropped out" or deactivated, meaning their outputs are set to zero. This prevents the network from relying too heavily on any individual neuron and encourages the learning of more distributed and robust representations. Variants of dropout, such as DropConnect \cite{Wan2013RegularizationON}, extend this idea by randomly dropping connections between layers instead of entire neurons. Techniques like Stochastic Depth \cite{huang2016deep} and LayerDrop \cite{Fan2019ReducingTD} have also explored randomly dropping out entire layers during training. This can be seen as an extreme form of activation noise, where the noise is applied at the layer level.

	Our proposed method, VISP, builds upon this rich history of stochastic
	regularization by introducing a novel approach that adaptively controls the
	stochasticity based on the estimated volatility of the network’s parameters.
	Unlike existing methods that often use fixed noise schedules or probabilities,
	VISP aims to dynamically adjust the regularization strength in a more fine-grained
	manner, informed by the training dynamics themselves.

	\section{VISP Description}

\begin{figure}[h]
		\centering
		\includegraphics[width=0.7 \textwidth]{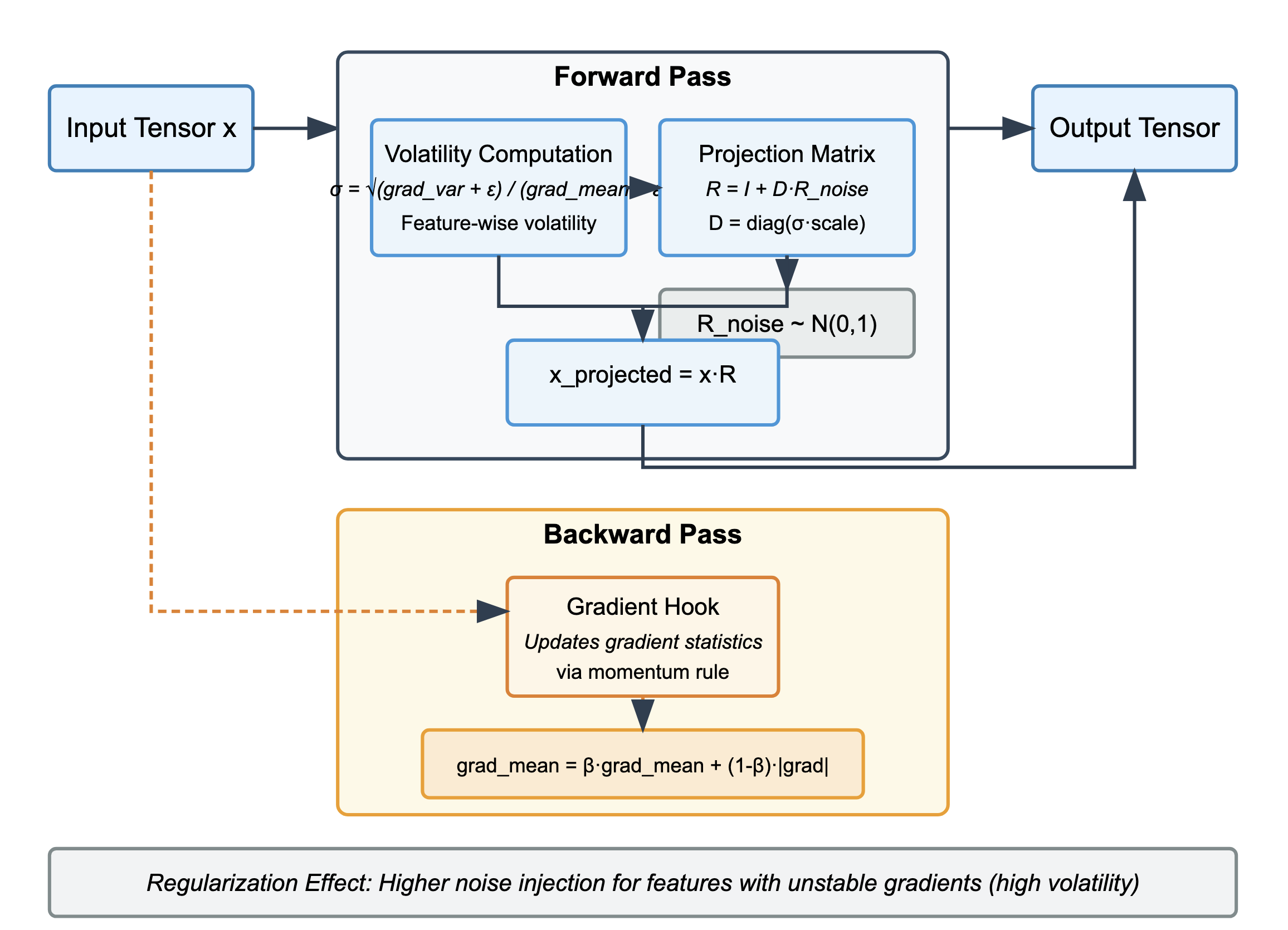}
		\caption{Overview of the calculations involved in a VISP layer.}
		\label{fig:visp_overview}
	\end{figure}

	In this section, we describe the Volatility Informed Stochastic Projection (VISP) mechanism. VISP adaptively regularizes the network by modulating noise injection based on gradient volatility. Figure \ref{fig:visp_overview} provides an overview of the calculations involved in a VISP layer. This includes computing the volatility of gradients, constructing a stochastic projection matrix, and applying this matrix to the activations during the forward pass.

	\subsection{Volatility Computation}

	Let \(\mathbf{x} \in \mathbb{R}^{B \times d}\) denote the input activations for a fully-connected layer (or \(\mathbf{x} \in \mathbb{R}^{B \times C \times H \times W}\) for convolutional layers, where \(C\) corresponds to \(d\) in this context). During backpropagation, we denote the gradient of the loss \(\mathcal{L}\) with respect to \(\mathbf{x}\) as \(\mathbf{g} \in \mathbb{R}^{B \times d}\). For each feature \(i \in \{1,\ldots,d\}\), we maintain running estimates of the absolute gradient mean and variance via an exponential moving average:

	\[
	\mu_i^{(t+1)} = \beta\,\mu_i^{(t)} + (1-\beta) \, \mathbb{E}_{\text{batch}} \left[|g_i|\right],
	\]
	\[
	\sigma_i^{2\,(t+1)} = \beta\,\sigma_i^{2\,(t)} + (1-\beta) \, \mathbb{E}_{\text{batch}} \left[\left(g_i - \mu_i^{(t)}\right)^2\right],
	\]
	where \(\beta \in [0,1)\) is a momentum hyperparameter. We set this to 0.9 and do not tune it further. Using these statistics, we define the per-feature \emph{volatility} as:
	\[
	v_i = \frac{\sqrt{\sigma_i^2} + \epsilon}{\mu_i + \epsilon},
	\]
	where \(\epsilon > 0\) is included to ensure numerical stability. To calibrate the magnitude of the noise injection, we introduce a scaling factor \(\alpha\) (referred to as \texttt{volatility\_scale} in our implementation, and set to 0.01) such that the transformed volatility is given by:
	\[
	\tilde{v}_i = \alpha\, v_i.
	\]
	This volatility metric captures the relative variability of the gradient for each feature, thereby providing a data-driven signal to guide the regularization strength.

	\subsection{Stochastic Projection Matrix Construction}

	Given the transformed volatility, we construct a diagonal scaling matrix \(\mathbf{D} \in \mathbb{R}^{d \times d}\) defined as:
	\[
	\mathbf{D} = \operatorname{diag}(\tilde{v}_1, \tilde{v}_2, \ldots, \tilde{v}_d).
	\]
	Next, we sample a random noise matrix \(\mathbf{R}_{\text{noise}} \in \mathbb{R}^{d \times d}\) with entries drawn independently from a standard normal distribution, i.e., \(\left[\mathbf{R}_{\text{noise}}\right]_{ij} \sim \mathcal{N}(0,1)\). The stochastic projection matrix is then constructed as:
	\[
	\mathbf{R} = \mathbf{I}_d + \mathbf{D} \, \mathbf{R}_{\text{noise}},
	\]
	where \(\mathbf{I}_d\) is the \(d \times d\) identity matrix. The final projection is then applied to the input activations as:
	\[
	\mathbf{x}' = \mathbf{x}\,\mathbf{R}.
	\]
	In the convolutional setting, a similar procedure is applied by reshaping the activations from \(\mathbb{R}^{B \times C \times H \times W}\) to \(\mathbb{R}^{(B \cdot H \cdot W) \times C}\), performing the projection along the channel dimension, and then reshaping the result back to its original dimensions.

	\subsection{Integration with Training Process}

	VISP is applied dynamically during the forward pass in training. The stochastic projection replaces the standard identity mapping in the affected layers, ensuring that each forward pass incorporates a unique, data-dependent perturbation. Importantly, the projection is only applied during training; during evaluation, VISP acts as an identity mapping to preserve the learned representations.

	To update the gradient statistics, a backward hook is attached to the projected activations. Let \(\mathbf{g}\) denote the gradient with respect to \(\mathbf{x}'\). The hook function updates the running averages as:
	\[
	\mu^{(t+1)} \leftarrow \beta\, \mu^{(t)} + (1-\beta) \, \mathbb{E}[|g|],
	\]
	\[
	\sigma^{2\,(t+1)} \leftarrow \beta\, \sigma^{2\,(t)} + (1-\beta) \, \mathbb{E}\left[\left(g - \mu\right)^2\right],
	\]
	where the expectation is computed over the batch (and spatial dimensions for convolutional layers). These updated statistics are then used in subsequent forward passes to adjust the stochastic projection matrix, ensuring that the regularization adapts to the evolving dynamics of the network.

	In summary, VISP leverages gradient volatility to modulate a stochastic projection applied to layer activations. By coupling adaptive noise injection with continuous gradient statistics updates, VISP regularizes the network in a data-dependent manner that encourages the learning of robust representations without unduly distorting the feature space.

	\subsection{Inference Process}

	During inference, it is essential that the model’s behavior is deterministic and that the learned representations remain unperturbed. Therefore, the VISP module is designed to be active only during training. In evaluation mode, the VISP module bypasses the stochastic projection and directly returns the input activations. Mathematically, for any input \(\mathbf{x}\), the output of the VISP module during inference is given by:
	\[
	\mathbf{x}' = \mathbf{x}.
	\]
	This design ensures that the adaptive noise injection, which is beneficial for regularization during training, does not interfere with the model’s predictions during testing. By disabling the VISP transformation at inference time, we guarantee that the stochasticity introduced during training does not affect the final output, thereby preserving the integrity of the learned representations.

	\section{Experiments}

	In this section, we conduct extensive experiments across several benchmark datasets, including MNIST, CIFAR-10, and SVHN, to demonstrate the effectiveness of VISP.

	\subsection{MNIST}

	First, we evaluate the effectiveness of the proposed VISP mechanism on the MNIST handwritten digit dataset. MNIST consists of 60,000 training images and 10,000 test images, each of size \(28 \times 28\) pixels in grayscale. For our experiments, all images are normalized to have zero mean and unit variance.

	We compare three variants of a fully-connected network. The baseline model, referred to as “Standard (No VISP)”, is a multilayer perceptron composed of three fully-connected layers with ReLU activations without the VISP layer. In contrast, the “Fixed Noise” variant uses the identical architecture as the baseline; however, it replaces the adaptive VISP module with a fixed noise projection that applies a constant scaling factor, independent of gradient volatility. Specifically, in the Fixed Noise model, during training, Gaussian noise with a standard deviation of 0.2 is added to the input, and a noise standard deviation of 0.5 is applied to the activations of the hidden layers. Finally, our proposed model, VISP, incorporates the Volatility Informed Stochastic Projection module at three key locations—before the first hidden layer, between the first and second hidden layers, and before the final classification layer—where it computes gradient volatility to adaptively scale a stochastic projection matrix.

	Each network consists of an input layer mapping the flattened \(28\times28\) images to 784 features, two hidden layers with 1024 neurons each, and an output layer with 10 neurons corresponding to the digit classes. All models are trained for 200 epochs using a mini-batch size of 512. The optimizer used is stochastic gradient descent (SGD) with a learning rate of 0.01 and momentum of 0.9.

	\begin{figure}[h]
		\centering
		\includegraphics[width=1.0 \textwidth]{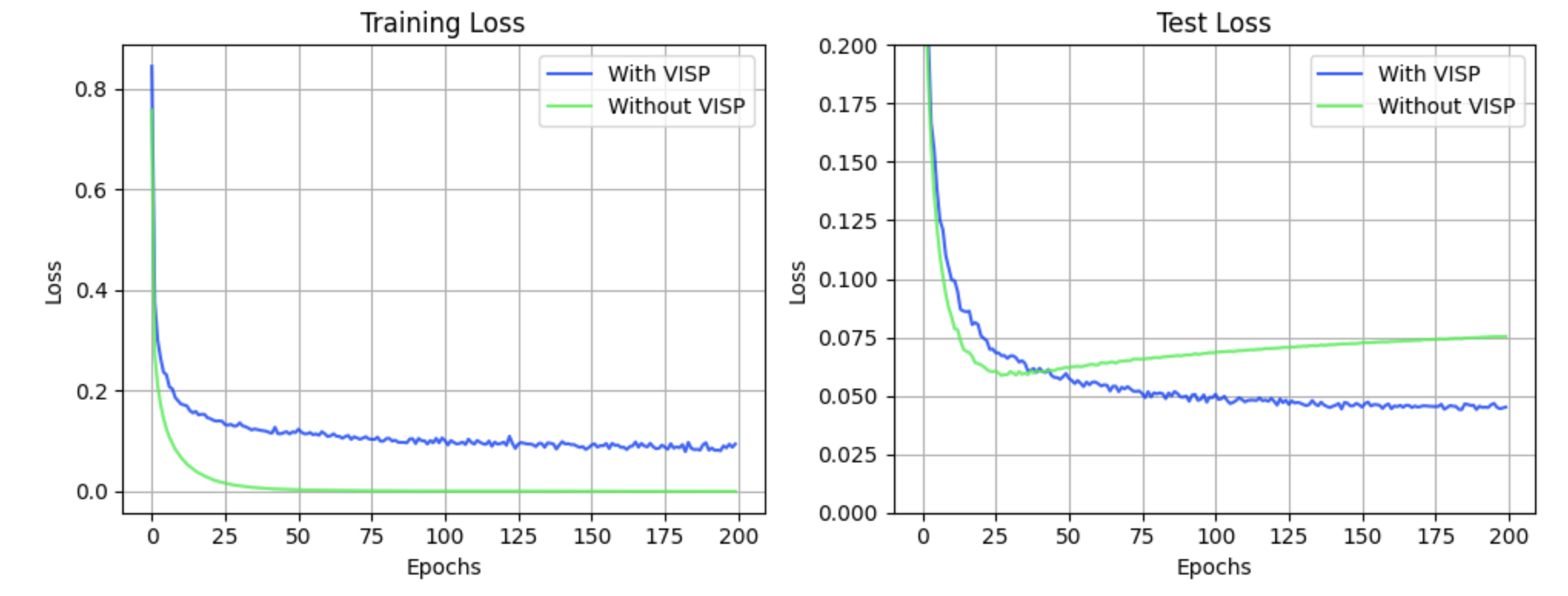}
		\caption{Training and test loss curves on MNIST with and without VISP layer. While the baseline model achieves lower training loss, its test loss rises sharply, indicating overfitting. In contrast, VISP maintains a slightly higher training loss but converges to a substantially lower test loss, demonstrating improved generalization through adaptive noise injection based on gradient volatility.}
		\label{fig:mnist_loss}
	\end{figure}

	\begin{figure}[h]
		\centering
		\includegraphics[width=0.5 \textwidth]{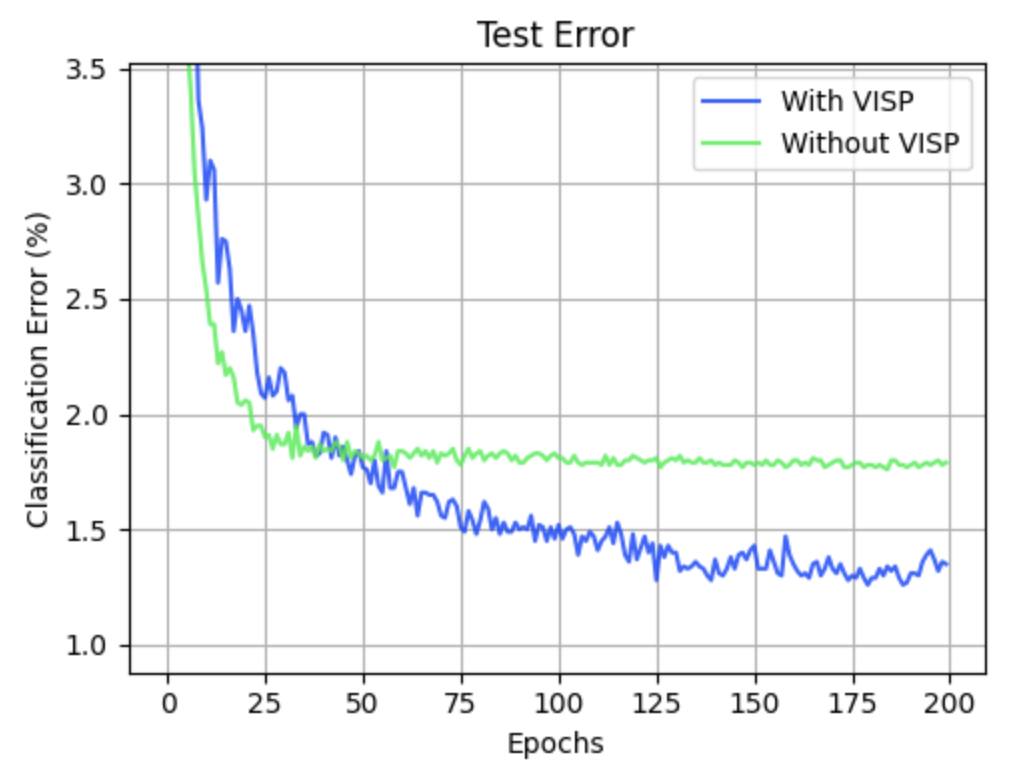}
		\caption{Test error curves on MNIST with and without VISP layer.}
		\label{fig:mnist_error}
	\end{figure}

	Figure~\ref{fig:mnist_loss} illustrates the evolution of the training and test loss over 200 epochs for MNIST, comparing the baseline model (without VISP) against the same architecture augmented with VISP. Although the baseline achieves a lower training loss, it rapidly begins to overfit, as indicated by its rising test loss. In contrast, the VISP model maintains a slightly higher training loss—reflecting the effect of adaptive stochastic projection—but converges to a significantly lower test loss. This behavior underscores the core benefit of VISP: by injecting noise selectively based on gradient volatility, it curbs overfitting more effectively than the non-regularized baseline, leading to better generalization performance on unseen data. 
	
	Similarly, Figure~\ref{fig:mnist_error} depicts the test error rates over the same 200 epochs for the baseline and VISP models. Both approaches initially exhibit a rapid decrease in error, but the VISP model continues to refine its predictions, ultimately converging to a lower error rate. In contrast, the baseline saturates at a higher error level, reflecting its tendency to overfit the training data. These findings reinforce the notion that VISP’s adaptive stochastic projection provides a more robust form of regularization, leading to consistently improved generalization performance.

	\begin{table}[ht]
	\centering
	\caption{Test error on MNIST for models without VISP, with fixed noise, and with VISP.}
	\begin{tabular}{l c}
	\toprule
	\textbf{Technique} & \textbf{Test Error (\%)} \\
	\midrule
	Standard (No VISP) & 1.77 \\
	Fixed Noise        & 1.44 \\
	VISP               & 1.28 \\
	\bottomrule
	\end{tabular}
	\label{tab:mnist_results}
	\end{table}

	Table~\ref{tab:mnist_results} summarizes the test error rates. As shown in the table, the baseline model without VISP achieves a final test error of 1.77\%, whereas adding a fixed noise projection reduces the error to 1.44\%. Our proposed VISP method yields the lowest test error at 1.28\%, demonstrating the benefit of tailoring stochastic projection using network’s evolving gradient statistics. Taken together, these results highlight the effectiveness of VISP as an adaptive regularization strategy on MNIST, achieving superior generalization compared to both an unregularized baseline and a fixed-noise alternative.

	\subsection{CIFAR-10}

	We further evaluate VISP on CIFAR-10, a widely used benchmark dataset consisting of 50,000 training images and 10,000 test images, each a \(32 \times 32\) color image from 10 object classes. All images are normalized channel-wise to zero mean and unit variance.

	We adopt a convolutional neural network architecture comprising three convolutional blocks followed by a fully-connected classifier. In the first block, a \(3 \times 3\) convolution maps the input (with three channels for CIFAR-10) to 64 feature maps, which are then passed through a ReLU activation and a VISP module (if applied) before a max pooling layer halves the spatial dimensions. In the second block, a \(3 \times 3\) convolution increases the channel depth from 64 to 128, followed by ReLU, another VISP module (if applied), and additional max pooling. The third block employs a \(3 \times 3\) convolution to further raise the number of channels from 128 to 256, again followed by ReLU and a VISP module if applied. Global average pooling is used to reduce the spatial dimensions to \(1 \times 1\), and the resulting feature vector is fed into a fully-connected layer that produces predictions for 10 classes. Similar to our MNIST experiments, we compare three variants of this network: a baseline model without any additional noise injection, a “Fixed Noise” model in which a constant noise scaling with a noise standard deviation of 0.05 is applied to the activations, and the proposed variant with stochastic projection based on gradient volatility. We train each model for 200 epochs with stochastic gradient descent (SGD), using a batch size of 512, a learning rate of 0.01, momentum of 0.9.

	\begin{figure}[h]
		\centering
		\includegraphics[width=0.5 \textwidth]{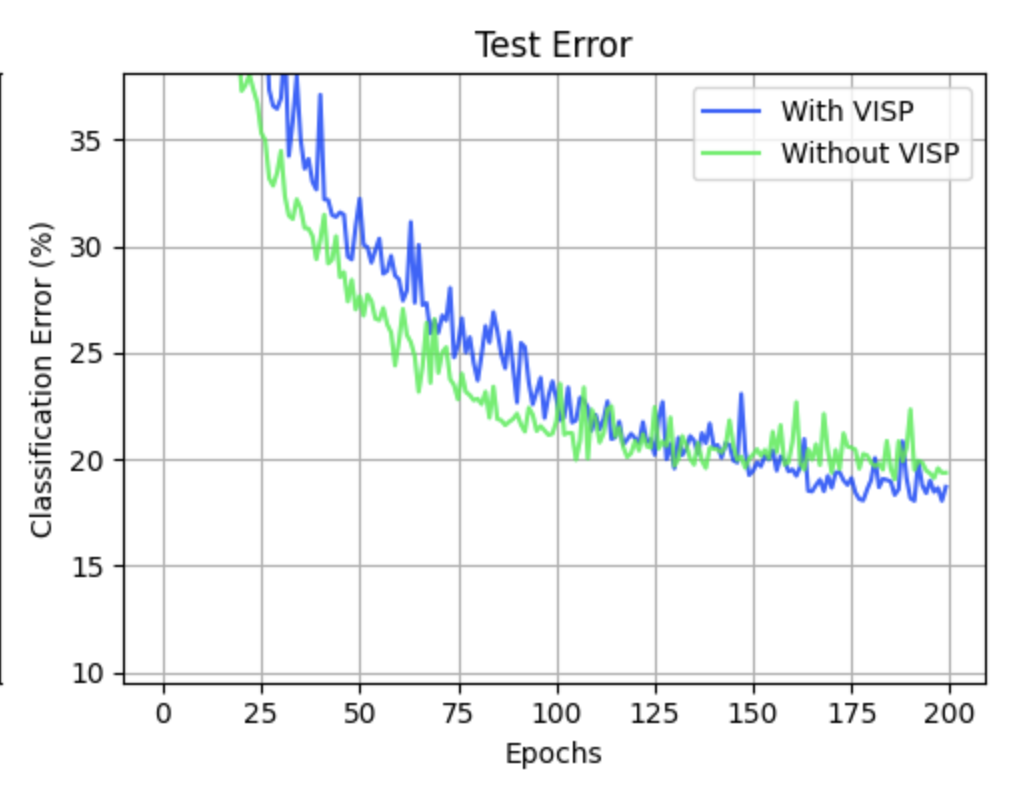}
		\caption{Test error curves on CIFAR-10 with and without VISP layer.}
		\label{fig:cifar10_error}
	\end{figure}

	Figure~\ref{fig:cifar10_error} shows the evolution of the test error for the models with and without VISP over 200 epochs. The VISP model continues to improve and converges to a lower final test error. Quantitatively, Table~\ref{tab:cifar10_results} summarizes the final test error values. The baseline (No VISP) model reaches 19.05\% error, while Fixed Noise slightly reduces it to 18.91\%. In contrast, VISP achieves a test error of 18.05\%, indicating a clear advantage from adaptively scaling the stochastic projection based on gradient volatility.

	\begin{table}[ht]
	\centering
	\caption{Final test error on CIFAR-10 for baseline, fixed noise, and VISP models.}
	\label{tab:cifar10_results}
	\begin{tabular}{l c}
	\toprule
	\textbf{Technique} & \textbf{Test Error (\%)} \\
	\midrule
	Standard (No VISP) & 19.05 \\
	Fixed Noise        & 18.91 \\
	VISP               & 18.05 \\
	\bottomrule
	\end{tabular}
	\end{table}

	These results confirm that VISP provides a more effective form of regularization than both the unregularized baseline and a naive fixed noise approach. By leveraging gradient volatility, VISP tailors the noise injection to the evolving state of the network, mitigating overfitting and promoting more robust feature representations. This performance gap is consistent with the MNIST findings and underscores the general applicability of VISP to different architectures and datasets.

	\subsection{SVHN}

	The Street View House Numbers (SVHN) dataset presents a real-world challenge for digit recognition, consisting of over 73,000 training images and roughly 26,000 test images captured from street scenes. Unlike more curated datasets, SVHN contains significant variability due to differences in lighting, background clutter, and scale, making it an ideal testbed for evaluating robust regularization methods.

	\begin{figure}[h]
		\centering
		\includegraphics[width=0.5 \textwidth]{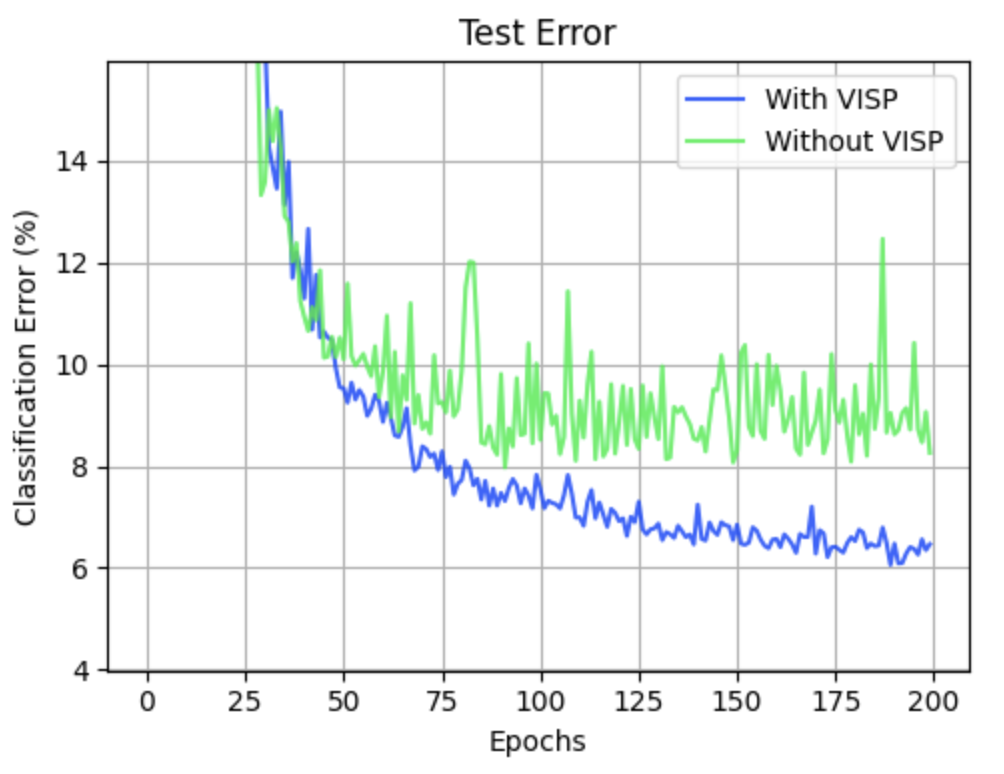}
		\caption{Test error curves on SVHN with and without VISP layer.}
		\label{fig:svhn_error}
	\end{figure}

	In our SVHN experiments, we also evaluate the performance of three network configurations: a baseline model without any additional regularization, a variant employing fixed noise injection, and our proposed VISP model that adapts noise injection based on gradient volatility. The final test performance is reflected by the test error plot (see Figure~\ref{fig:svhn_error}) and the summarized results in Table~\ref{tab:svhn_results}.

	\begin{table}[ht]
	\centering
	\caption{Test error on SVHN for models without VISP, with fixed noise, and with VISP.}
	\label{tab:svhn_results}
	\begin{tabular}{l c}
	\toprule
	\textbf{Technique} & \textbf{Test Error (\%)} \\
	\midrule
	Standard (No VISP)     & 8.25 \\
	Fixed Noise & 8.39 \\
	VISP        & 6.21 \\
	\bottomrule
	\end{tabular}
	\end{table}

	Our findings reveal that the VISP-augmented model achieves a significant reduction in test error, attaining 6.21\%, compared to 8.25\% for the baseline and 8.39\% for the fixed noise variant. This improvement suggests that the adaptive regularization provided by VISP is particularly effective under the heterogeneous conditions of SVHN, where the dynamic adjustment of noise injection better captures the underlying variability in the data. The marked improvement with VISP indicates that leveraging gradient volatility to modulate noise injection can be especially beneficial in scenarios with high data variability, leading to enhanced generalization performance on challenging, real-world datasets like SVHN.

	\section{Analysis of VISP Dynamics}

	Next, a series of diagnostic plots—ranging from the evolution of gradient volatility and the spectral properties of the projection matrix to detailed activation distribution comparisons—provides deep insights into how VISP modulates the network’s internal representations. The plots are from VISP-based MNIST training.

	\subsection{Volatility Evolution}

\begin{figure}[h]
		\centering
		\includegraphics[width=1.0 \textwidth]{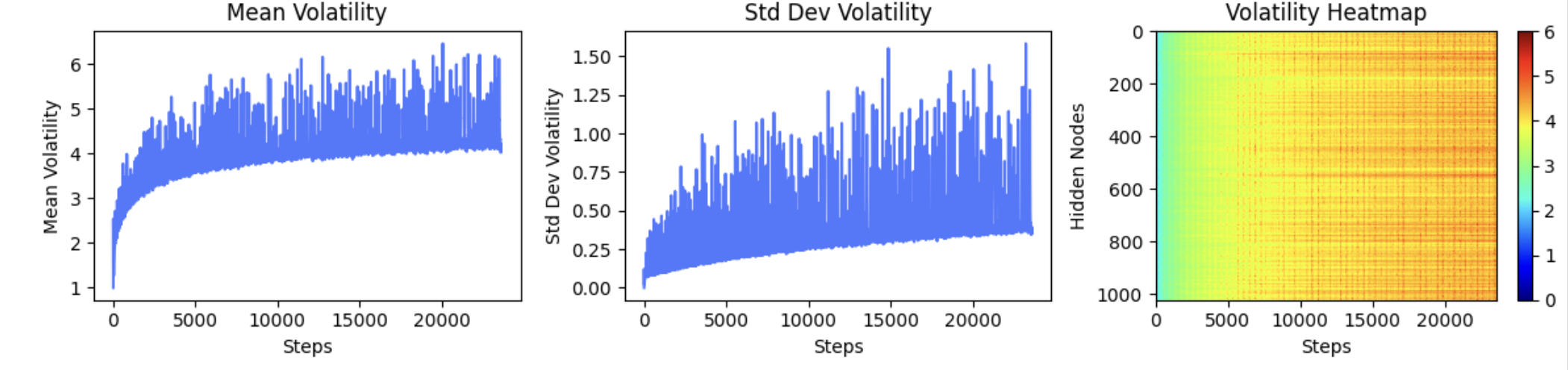}
		\caption{Evolution of gradient volatility in a VISP-augmented network over training steps. The left panel shows the mean volatility, while the center panel depicts the standard deviation, and the right panel is a heatmap of per-neuron volatility.}
		\label{fig:volatility_evolution}
	\end{figure}

	Figure~\ref{fig:volatility_evolution} presents three perspectives on how volatility unfolds within a VISP-augmented network during training. The left plot shows the mean volatility across all neurons at each step, the center plot depicts the standard deviation of volatility values, and the right plot provides a heatmap of per-neuron volatility over time. They are from the first hidden layer. Interpreting these plots offers insight into the adaptive nature of VISP and clarifies why such volatility dynamics can be beneficial.

	\paragraph{\textbf{Rising Mean Volatility.}}

	The gradual increase in the mean volatility (left plot) indicates that, as training progresses, the network begins to identify a broader set of activations that exhibit higher gradient fluctuations. During the early stages of training, the gradients are typically small and might be relatively uniform, resulting in low volatility. As training progresses, however, the network starts to specialize: certain neurons may become more responsive to particular features and thus exhibit larger gradient magnitudes, while the variability (variance) in those gradients can also increase. In such cases, even if the network is converging and overall error is decreasing, the ratio that defines volatility can increase if the relative dispersion of the gradients grows faster than their average magnitude. In essence, an increase in volatility suggests that the network is identifying and emphasizing specific features—those with higher variability—which is exactly what VISP is designed to regularize. From a VISP standpoint, this behavior is expected and advantageous: activations with larger volatility receive stronger stochastic projection, preventing them from overfitting and encouraging more robust representations.

	\paragraph{\textbf{Growing Spread in Volatility.}}
	The center plot shows that the standard deviation of the computed volatility increases as training progresses, indicating a widening gap between neurons with highly volatile activations and those with more stable responses. Early in training, when the model has yet to identify which neurons are most critical, the volatility standard deviation is low. However, as learning advances, certain neurons exhibit much higher gradient fluctuations than others. This differentiation results in a broader spread of volatility values, as reflected by the rising standard deviation in the center plot. Crucially, VISP leverages this widening gap by adaptively scaling stochastic projection: neurons with higher volatility receive stronger regularization, while those with lower volatility are perturbed less. 

	\paragraph{\textbf{Heatmap of Per-Neuron Volatility.}}
	The heatmap (right plot) provides a neuron-by-neuron view of how volatility evolves. Early in training, the colors are relatively uniform, indicating that no single neuron stands out by exhibiting high gradient fluctuations. Over time, certain rows (neurons) transition toward higher values, forming bands of heightened volatility.  In other words, some neurons continue to exhibit relatively low gradient fluctuations, while others become distinctly more volatile. This divergence is precisely what VISP is designed to exploit—neurons with high volatility are adaptively regularized more aggressively, whereas neurons with lower volatility remain relatively unperturbed. This pattern aligns with the idea that, as the model discovers specific features that contribute disproportionately to the training loss, these features’ gradients exhibit greater variability—prompting VISP to intensify the regularization in exactly those directions. Such selective noise injection helps the network devote its capacity to learning features that are genuinely informative, rather than being swamped by uniform noise across all neurons.

	\paragraph{\textbf{Implications for Generalization.}}
	From a broader perspective, these volatility trends are a \emph{positive} sign that VISP is operating as intended. Rather than applying uniform noise, the method adapts to the network’s internal dynamics by focusing regularization where it is most needed. This targeted approach can mitigate overfitting more effectively than a one-size-fits-all noise injection scheme. The result, as evidenced in our empirical results, is improved test performance without sacrificing the representational flexibility of more stable parts of the network.

	In summary, the rising mean, widening spread, and neuron-specific patterns in volatility collectively underscore VISP’s adaptive nature. By zeroing in on the activations that exhibit high gradient volatility, VISP ensures that the network remains robust in precisely those areas most prone to overfitting, leading to stronger generalization outcomes.

	\subsection{Spectral Dynamics of the Projection Matrix}

	To further understand the impact of VISP on the network's internal representations, we analyze the spectral properties of the stochastic projection matrix \(\mathbf{R}\) over the course of training. 
	
	\begin{figure}[h]
		\centering
		\includegraphics[width=0.5 \textwidth]{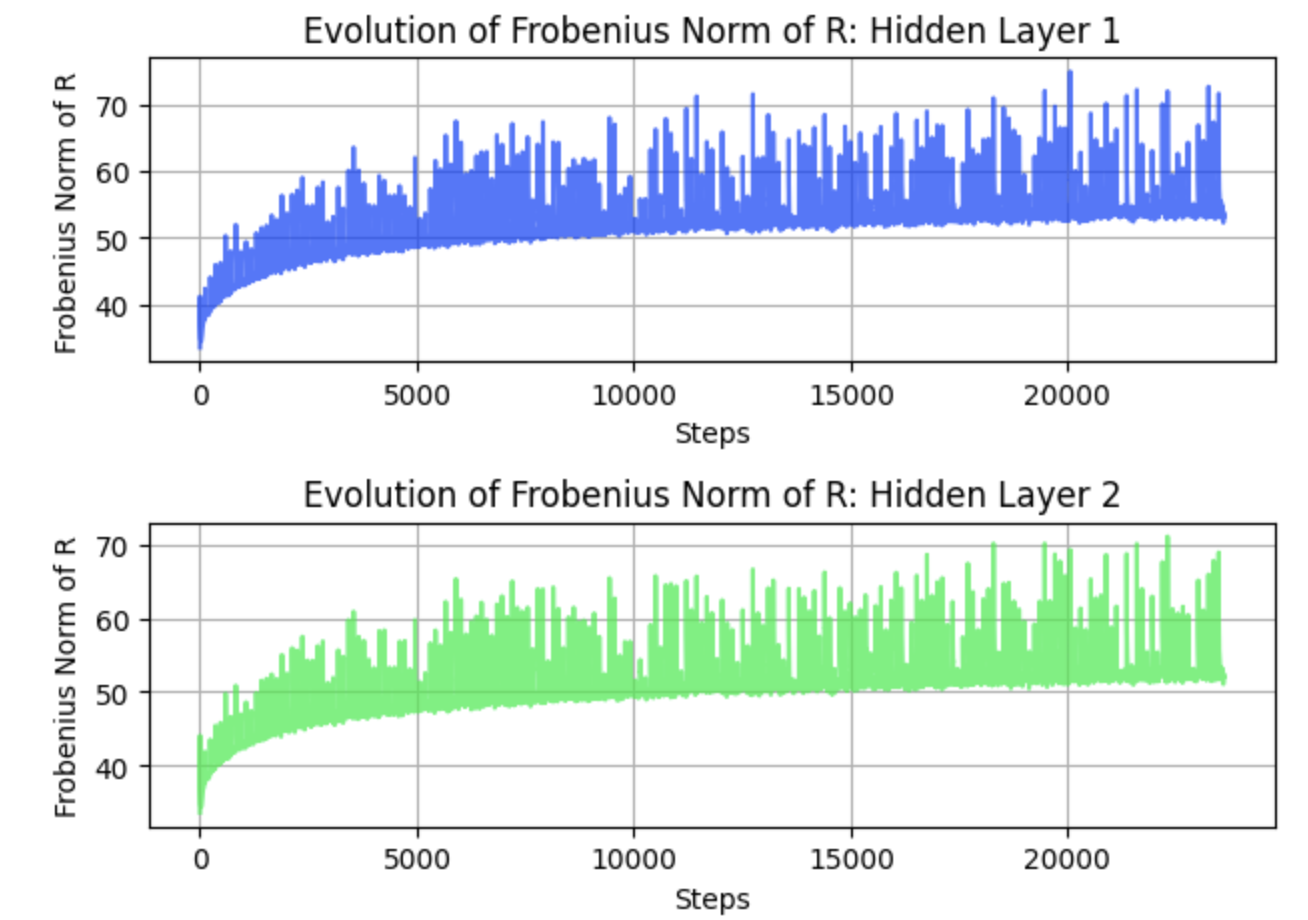}
		\caption{Evolution of the Frobenius norm of the projection matrix \(R\) over training steps.}
		\label{fig:f_norm_evolution}
	\end{figure}

	Figure~\ref{fig:f_norm_evolution} plots the Frobenius norm of the projection matrix \(\mathbf{R}\) for the first (top plot) and second (bottom plot) hidden layers over training steps. In both layers, the norm rises from an initial lower bound toward a higher plateau, indicating that \(\mathbf{R}\) increasingly diverges from the identity matrix. This divergence reflects the growing influence of adaptive stochastic projection: as gradient volatility accumulates, the diagonal scaling \(\mathbf{D}\) in VISP scales the random perturbations more aggressively for the activations identified as volatile. Notably, the Frobenius norm does not explode to arbitrarily large values, suggesting that VISP remains stable and does not overwhelm the network with excessive noise.

	\begin{figure}[h]
		\centering
		\includegraphics[width=0.5 \textwidth]{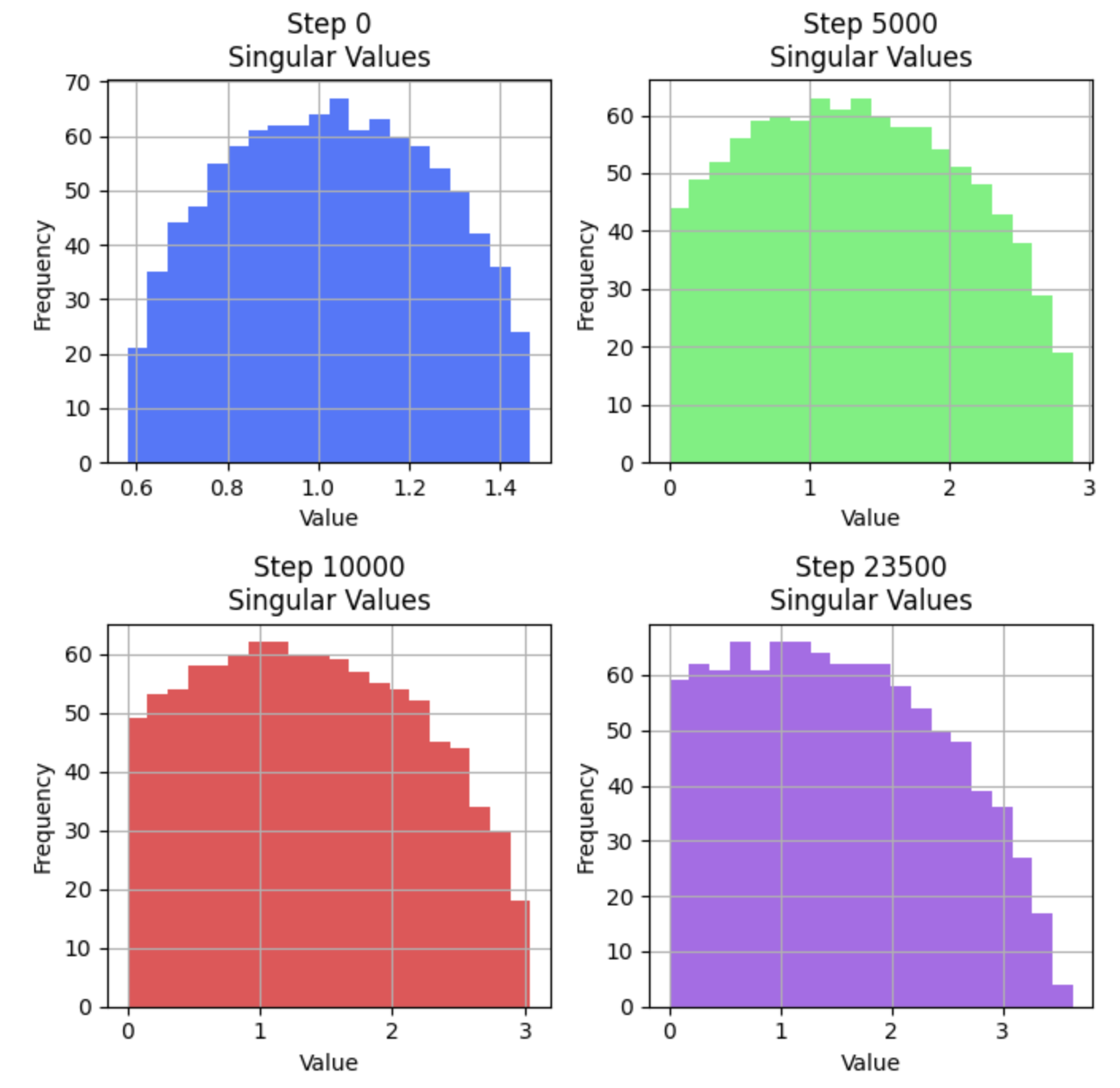}
		\caption{Evolution of the singular value distribution of the projection matrix \(R\) over four different training steps.}
		\label{fig:singular_values}
	\end{figure}

	Figure~\ref{fig:singular_values} shows histograms of the singular values of \(\mathbf{R}\) at four training steps (0, 5000, 10{,}000, and 23{,}500). At step 0, the distribution is relatively narrow and clustered near 1, indicating that \(\mathbf{R}\) starts close to an identity-like transform. As training proceeds, the distribution broadens and shifts rightward, with some singular values exceeding 3. This widening range implies that \(\mathbf{R}\) becomes more anisotropic, selectively amplifying certain directions in the feature space while damping others. From a regularization standpoint, such anisotropy is desirable: activations with higher volatility receive stronger perturbations, thereby mitigating overfitting, while more stable activations remain closer to their original representation. 

	Taken together, these spectral dynamics confirm that VISP’s projection matrix evolves in a controlled yet non-uniform manner. Rather than uniformly scattering activations, it directs its noise injection toward the most volatile activations. The absence of singular values collapsing toward zero or blowing up to extreme magnitudes underscores the stability of the method. By adaptively modulating the projection matrix in response to gradient volatility, VISP provides a principled mechanism for targeted regularization that balances feature preservation with the need for robust noise injection.

	\subsection{Activation Distribution}

\begin{figure}[h]
		\centering
		\includegraphics[width=0.6 \textwidth]{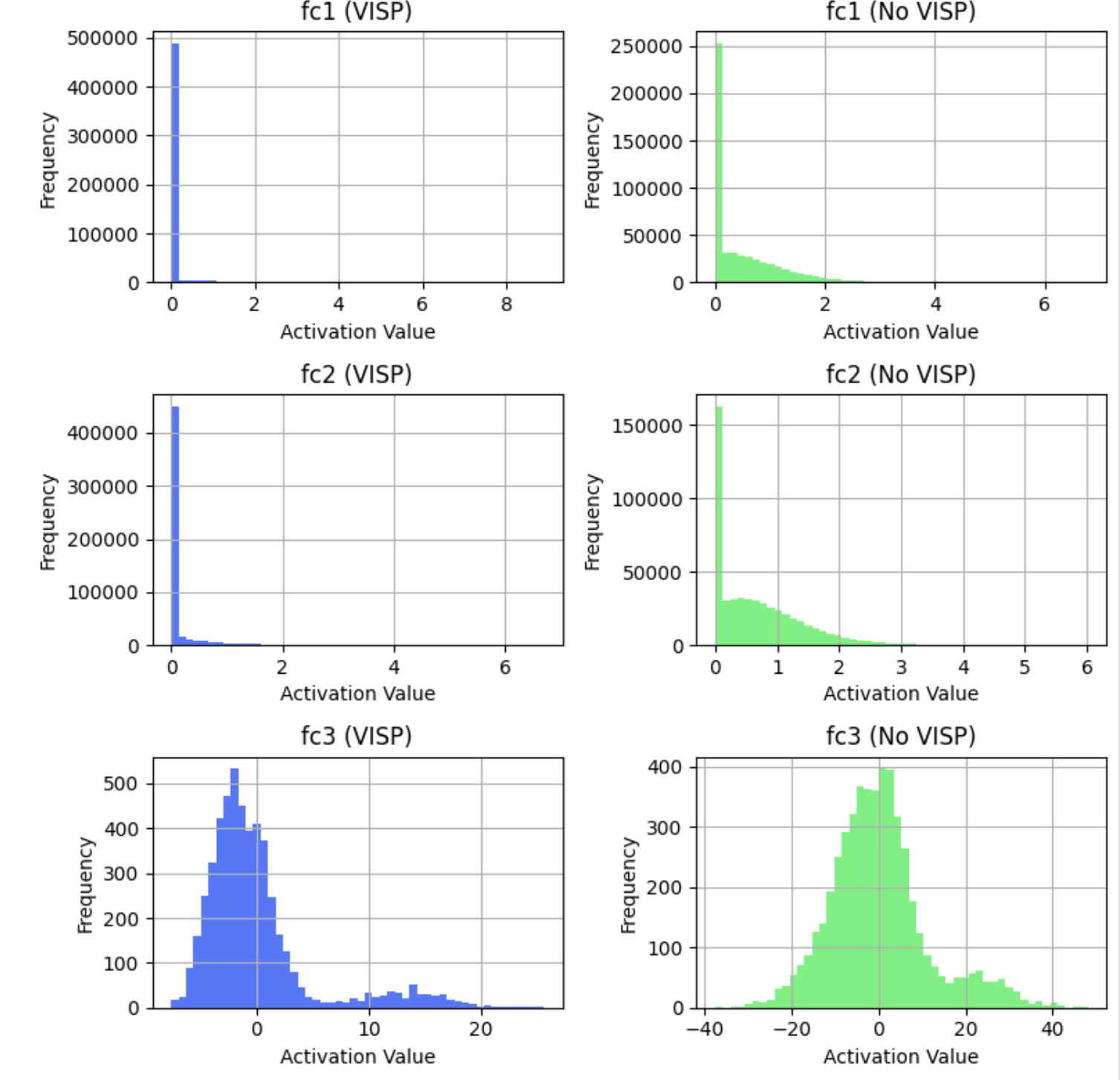}
		\caption{Activation distributions for selected layers in a network trained with VISP compared to a baseline.}
		\label{fig:visp_activation_distribution}
	\end{figure}

	To gain further insights into how VISP affects the network's internal representations, we compare the activation distributions of networks trained with and without VISP. 	Figure~\ref{fig:visp_activation_distribution} compares the activation histograms of three fully-connected layers (fc1, fc2, fc3) in a network trained with VISP (left column) against an otherwise identical baseline network without additional regularization (right column). In each layer, the VISP distributions are visibly narrower, suggesting that the adaptive stochastic projection moderates large activations while preserving stable ones. For instance, in fc3, the baseline exhibits activations ranging from approximately \(-40\) to \(40\), whereas the VISP distribution remains within a considerably smaller interval. This contrast implies that, by selectively injecting stochasticity based on gradient volatility, VISP prevents certain neurons from growing disproportionately large or negative, thereby mitigating potential overfitting. The overall effect is a more balanced activation profile in each layer, consistent with the improved generalization results observed in earlier sections.

	These distributions suggest that the adaptive stochastic projection in VISP regularizes through gradient volatility. As training progresses, neurons identified as “volatile” by VISP receive proportionally stronger perturbations, curbing their ability to grow too large. Meanwhile, neurons deemed more stable remain closer to their original distributions, preserving essential signal. This selective regularization not only helps control the activation range but also reduces the risk of overfitting by ensuring that no single neuron or subset of neurons dominates the representation with excessively large outputs. As such, the VISP model’s activation histograms remain comparatively narrower and more centered, aligning with the improved generalization performance observed in our experiments.

	\section{Conclusion}

	In this work, we introduce VISP—Volatility Informed Stochastic Projection—a novel adaptive regularization strategy that leverages gradient volatility to modulate stochastic projection within deep neural networks. By dynamically computing per-feature (or per-channel) volatility and employing it to scale a stochastic projection matrix, VISP selectively targets volatile features and activations with stronger regularization. Our experimental evaluations on MNIST, CIFAR-10, and SVHN demonstrate that networks augmented with VISP achieve superior generalization performance compared to both unregularized baselines and fixed-noise variants.

	Beyond the performance gains, our in-depth analysis of VISP's internal dynamics—encompassing volatility evolution, spectral properties of the projection matrix, and activation distributions—provides critical insights into how data-dependent noise injection can stabilize training and mitigate overfitting. The observed evolution of these metrics confirms that VISP adapts its regularization strength in a principled manner, ensuring that the network remains robust while retaining its representational capacity.

	\bibliographystyle{splncs04}
	\bibliography{references.bib}
\end{document}